%% file: imdd_dfe.tex
\documentclass[letterpaper,10pt]{article} 

\usepackage{opticameet3} %

\usepackage{mathtools,amssymb,tikz,bm,siunitx, booktabs, subcaption}
\usepackage[breaklinks,colorlinks=true,bookmarks=false,citecolor=blue,urlcolor=blue]{hyperref} %
\usetikzlibrary{positioning, shapes.geometric}            %

\input{kit_colors.tex}

\newcommand{\argmax}{\text{arg}\,\text{max}}
\usepackage{wrapfig}
\usepackage[font={footnotesize}]{caption}           %
\DeclareMathAlphabet{\mathcal}{OMS}{cmsy}{m}{n}     %

\newcommand\blfootnote[1]{%
  \begingroup
  \renewcommand\thefootnote{}\footnote{#1}%
  \addtocounter{footnote}{-1}%
  \endgroup
}

\usepackage{pgfplots}
\usetikzlibrary{arrows,calc,fit,matrix,positioning,shapes,shadows,trees,mindmap,tikzmark,arrows.meta,angles,quotes,babel,spy}

\usetikzlibrary{external}

\begin{document}

\title{Spiking Neural Network Decision Feedback Equalization for IM/DD Systems}

\author{Alexander von Bank, Eike-Manuel Edelmann, and Laurent Schmalen}

\address{\mbox{Communications Engineering Lab, Karlsruhe Institute of Technology, 76187 Karlsruhe, Germany}}
\email{\texttt{alexander.bank@kit.edu, edelmann@kit.edu}} %

\vspace{-3mm}
\begin{abstract}
A spiking neural network (SNN) equalizer with a decision feedback structure is applied to an IM/DD link with various parameters.
The SNN outperforms linear and artificial neural network (ANN) based equalizers.
\vspace{-0mm}
\end{abstract}

\blfootnote{This work has received funding from the European Research Council (ERC) under the European Union’s Horizon 2020 research and innovation programme (grant agreement No. 101001899).
Parts of this work were carried out in the framework of the CELTIC-NEXT project AI-NET-ANTILLAS (C2019/3-3) funded by the German Federal Ministry of Education and Research (BMBF) (grant agreement 16KIS1316).}

\vspace{-1mm}
\input{chapters/introduction.tex}

\input{chapters/SNN.tex}
\input{chapters/results.tex}

\end{document}

%% file: kit_colors.tex
\definecolor{kit-green100}{rgb}{0,.59,.51}
\definecolor{kit-green70}{rgb}{.3,.71,.65}
\definecolor{kit-green50}{rgb}{.50,.79,.75}
\definecolor{kit-green30}{rgb}{.69,.87,.85}
\definecolor{kit-green15}{rgb}{.85,.93,.93}
\definecolor{KITgreen}{rgb}{0,.59,.51}

\definecolor{KITpalegreen}{RGB}{130,190,60}
\colorlet{kit-maigreen100}{KITpalegreen}
\colorlet{kit-maigreen70}{KITpalegreen!70}
\colorlet{kit-maigreen50}{KITpalegreen!50}
\colorlet{kit-maigreen30}{KITpalegreen!30}
\colorlet{kit-maigreen15}{KITpalegreen!15}

\definecolor{KITblue}{rgb}{.27,.39,.66}
\definecolor{kit-blue100}{rgb}{.27,.39,.67}
\definecolor{kit-blue70}{rgb}{.49,.57,.76}
\definecolor{kit-blue50}{rgb}{.64,.69,.83}
\definecolor{kit-blue30}{rgb}{.78,.82,.9}
\definecolor{kit-blue15}{rgb}{.89,.91,.95}

\definecolor{KITyellow}{rgb}{.98,.89,0}
\definecolor{kit-yellow100}{cmyk}{0,.05,1,0}
\definecolor{kit-yellow70}{cmyk}{0,.035,.7,0}
\definecolor{kit-yellow50}{cmyk}{0,.025,.5,0}
\definecolor{kit-yellow30}{cmyk}{0,.015,.3,0}
\definecolor{kit-yellow15}{cmyk}{0,.0075,.15,0}

\definecolor{KITorange}{rgb}{.87,.60,.10}
\definecolor{kit-orange100}{cmyk}{0,.45,1,0}
\definecolor{kit-orange70}{cmyk}{0,.315,.7,0}
\definecolor{kit-orange50}{cmyk}{0,.225,.5,0}
\definecolor{kit-orange30}{cmyk}{0,.135,.3,0}
\definecolor{kit-orange15}{cmyk}{0,.0675,.15,0}

\definecolor{KITred}{rgb}{.63,.13,.13}
\definecolor{kit-red100}{cmyk}{.25,1,1,0}
\definecolor{kit-red70}{cmyk}{.175,.7,.7,0}
\definecolor{kit-red50}{cmyk}{.125,.5,.5,0}
\definecolor{kit-red30}{cmyk}{.075,.3,.3,0}
\definecolor{kit-red15}{cmyk}{.0375,.15,.15,0}

\definecolor{KITpurple}{RGB}{160,0,120}
\colorlet{kit-purple100}{KITpurple}
\colorlet{kit-purple70}{KITpurple!70}
\colorlet{kit-purple50}{KITpurple!50}
\colorlet{kit-purple30}{KITpurple!30}
\colorlet{kit-purple15}{KITpurple!15}

\definecolor{KITcyanblue}{RGB}{80,170,230}
\colorlet{kit-cyanblue100}{KITcyanblue}
\colorlet{kit-cyanblue70}{KITcyanblue!70}
\colorlet{kit-cyanblue50}{KITcyanblue!50}
\colorlet{kit-cyanblue30}{KITcyanblue!30}
\colorlet{kit-cyanblue15}{KITcyanblue!15}

%% file: chapters/introduction.tex
\section{Introduction}
\vspace*{-2mm}
Machine-learning-based equalizers have proven to enhance system performance for non-linear optical channels.
However, the performance of most equalizers depends on their complexity, leading to power-hungry receivers when implemented on digital hardware.
Compared to conventional digital hardware, neuromorphic hardware can massively reduce energy consumption when solving the same tasks~\cite{Ferreira}.
Spiking neural networks (SNNs) implemented on neuromorphic hardware mimic the human brain's behavior and promise energy-efficient, low-latency processing~\cite{BPTT}.
In~\cite{Bansbach}, an SNN-based equalizer with a decision feedback structure (SNN-DFE) has been proposed for equalization and demapping based on future and currently received, and already decided symbols.
For different multipath scenarios, i.e., linear channels, the SNN-DFE performs similarly to the classical decision feedback equalizer (CDFE) and artificial neural network (ANN) based equalizers.
For a 4-fold pulse amplitude modulation (PAM4) transmitted over an intensity modulation / direct detection (IM/DD) link suffering from \textit{chromatic dispersion} (CD) and non-linear impairments,~\cite{arnold22neuro} proposes an SNN that estimates the transmit symbols based on received symbols without feedback, no-feedback-SNN (NF-SNN). 
In simulations and experiments using the neuromorphic BrainScaleS-2 system,~\cite{arnold22neuro} outperforms linear minimum mean square error (LMMSE) equalizers and draws level with ANN-based equalizers.  
However, the approach of ~\cite{arnold22neuro} lacks feedback on already decided decisions, which can enhance equalization.
In this paper, we apply the SNN-DFE of~\cite{Bansbach} to the channel model of~\cite{arnold22neuro}. 
By incorporating decisions, the SNN-DFE outperforms the approach proposed by~\cite{arnold22neuro}.
As chromatic dispersion grows, the SNN-DFE benefits more from the decision feedback structure.
The code is available at \url{https://github.com/kit-cel/snn-dfe}.
\vspace*{-1mm}

%% file: chapters/SNN.tex
\newcommand{\e}{\mathrm{e}}

\section{Spiking-Neural-Network-based Decision Feedback Equalizer and Demapper}
\vspace*{-2mm}
Like ANNs, SNNs are networks of interconnected neurons; however, SNNs represent information using pulses (spikes) instead of numbers.
SNN neurons have an internal membrane state~$v(t)$, which depends on the time-variant input spike trains $s_j(t) \in \{0,1\},\,j \in \mathbb{N}$, emitted by the $j$-th upstream connected neuron.
The spike trains $s_j(t)$ cause the synaptic current $i(t)$, which charges the neuron's state $v(t)$.
If $v(t)$ surpasses a predefined threshold~$v_\text{th}$, i.e., if $v(t)>v_\text{th}$, the neuron emits an output spike. 
Afterward, the neuron resets its state to $v(t)=v_\text{r}$.
A common neuron model is the leaky-integrate-and-fire (LIF) model. It is described by~\cite{BPTT} $\frac{\mathrm{d}v(t)}{\mathrm{d}t} = -\frac{(v(t)-v_\text{r})+i(t)}{\tau_\text{m}} + \Theta(v(t)-v_\text{th}) (v_\text{r}-v_\text{th})$ and $\frac{\mathrm{d}i(t)}{\mathrm{d}t} = -\frac{i(t)}{\tau_\text{s}} + \sum_j w_j s_j(t)$, 
where $\tau_\text{m}$ and $\tau_\text{s}$ are the time constants of $v(t)$ and $i(t)$, $\Theta(\cdot)$ denotes the Heaviside function, and $w_j \in \mathbb{R}$ are the input weights.
The PyTorch-based SNN deep learning library \texttt{Norse}~\cite{norse} provides the approximate solutions of the equations as well as the backpropagation through time algorithm (BPTT) with surrogate gradients \cite{BPTT}, which is used for updating the network's weights. \\
\vspace{-4mm}
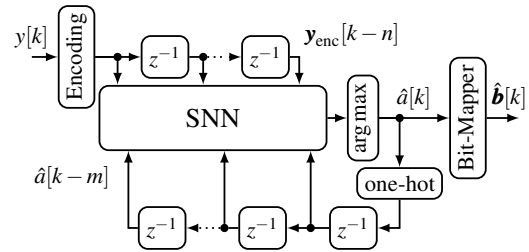
\begin{wrapfigure}{r}{0.45\textwidth}
    \vspace{-0.7cm}
    \begin{center}
        \resizebox{0.45\textwidth}{!}{
            \input{figures/dfe.tex}
        }
        \vspace*{-6mm}
        \caption{Proposed SNN-DFE structure}
        \label{fig:dfe}
    \end{center}
    \vspace{-1.1cm}
\end{wrapfigure}
Inspired by a DFE, we proposed an SNN-based equalizer and demapper (SNN-DFE) in~\cite{Bansbach}, whose architecture can be seen in Fig.~\ref{fig:dfe}.
The received symbol $y[k]$, where~$k$ denotes the discrete time, is translated into a spike signal using ternary encoding \cite{Bansbach} with $M_\text{t}=8$ input neurons per sample.
Based on the last $n=\big\lceil \frac{n_\text{tap}}{2}\big\rceil$ received symbols and the last $m=\big\lfloor \frac{n_\text{tap}}{2}\big\rfloor$ estimated symbols, the transmit symbol class estimate $\hat{a}[k]$ is obtained, where~$n_\text{tap}$ is the number of significant channel taps when sampling the channel with symbol rate.
Finally, a bit mapper translates the estimate $\hat{a}[k]$ to a corresponding bit sequence $\hat{\bm{b}}[k]$.

%% file: figures/dfe.tex
\begin{tikzpicture}[>=latex,thick]
    \def\zlen{0.65cm}
    \def\zdist{0.15cm}

    \node[draw,rectangle,rounded corners,minimum width=3.5cm, minimum height=1cm] (snn) {\large SNN};

    \node[above left=-0.37cm and 0.05cm of snn, draw, rectangle, rounded corners] (enc) {\rotatebox{90}{Encoding}};
    \node[left=0.4cm of enc] (y0) {};
    
    \node[draw, right=0.7cm of enc,rounded corners,minimum width=\zlen,minimum height=\zlen] (zy1) {$z^{-1}$};
    \node[draw, right=0.8cm of zy1,rounded corners,minimum width=\zlen,minimum height=\zlen] (zy2) {$z^{-1}$};
    \node[draw, right=\zdist+0.13cm of enc,circle, inner sep=1pt, fill=black] (nf1) {};
    \node[draw, right=\zdist of zy1,circle, inner sep=1pt, fill=black] (nf2) {};
    \node[above right=-3mm and 1mm of zy2] () {$\bm{y}_\text{enc}[k-n]$};

    \node[right=0.3cm of snn, draw, rectangle, rounded corners] (argmax) {\rotatebox{90}{$\argmax$}};
    \node[right=0.25cm of argmax,draw,circle, inner sep=1pt, fill=black] (n1) {};
    \node[right=0.7 of n1, draw, rectangle, rounded corners] (demap) {\rotatebox{90}{Bit-Mapper}};
    \node[right=0.5cm of demap] (xk) {};

    \node[draw,below=0.7cm of n1,rounded corners, rectangle] (onehot) {one-hot};
    \node[draw, below left=0.1cm and -0.33cm of onehot,rounded corners,minimum width=\zlen,minimum height=\zlen] (zx0) {$z^{-1}$};
    \node[draw, left =0.6cm+0.05cm of zx0,rounded corners,minimum width=\zlen,minimum height=\zlen] (zx1) {$z^{-1}$};
    \node[draw, left =0.8cm of zx1,rounded corners,minimum width=\zlen,minimum height=\zlen] (zx2) {$z^{-1}$};
    \node[draw, left=\zdist+0.07cm of zx0, circle, inner sep=1pt, fill=black] (nb1) {};
    \node[draw, left=\zdist of zx1, circle, inner sep=1pt, fill=black] (nb2) {};
    \node[below left=0.1cm and -0.3cm of snn] (xkn) {$\hat{a}[k-m]$};

    \def\yoff{0.43cm}
    \draw[->] (enc) -- (zy1);
    \draw[->] (enc) -- (nf1) -- +(0cm,-\yoff);
    \draw[->] (zy1) -- (nf2) -- +(0cm,-\yoff);
    \draw[dotted] (nf2) -- +(0.3cm,0cm);
    \draw[->] (nf2)+(0.4cm,0cm) -- (zy2);
    \draw[->] (zy2) -| +(0.52cm,-\yoff);
    \draw[->] (y0) -- (enc) node [pos=0.0,above] {$y[k]$};

    \def\xoff{1.2cm}
    \draw[->] (snn) -- (argmax);
    \draw[->] (argmax) -- (demap) node [midway, above] () {$\hat{a}[k]$};
    \draw[->] (demap) -- (xk) node [pos=0.7,above] {$\hat{\bm{b}}[k]$};
    \draw[->] (argmax) -| (onehot);
    \draw[->] (onehot) |- (zx0);

    \draw[->] (zx0) -- (zx1);
    \draw[->] (nb1) -- +(0cm,\xoff);
    \draw[->] (zx1) -- (nb2) -- +(0cm,\xoff);
    \draw[dotted] (nb2) -- +(-0.5cm,0cm);
    \draw[->] (nb2)+(-0.5cm,0cm) -- (zx2);
    \draw[->] (zx2) -| +(-0.52cm,\xoff);

\end{tikzpicture}

%% file: chapters/results.tex
\section{Results and Conclusion}
\vspace*{-4mm}
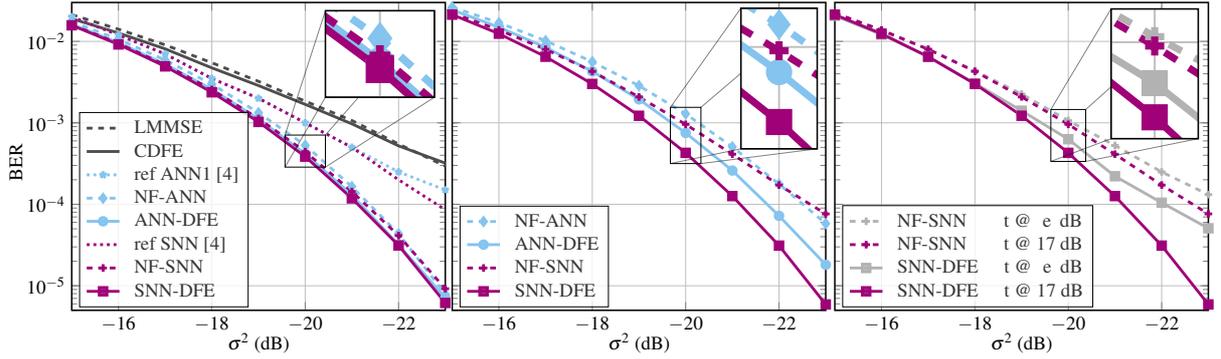
\begin{figure}[!htb]
    \hspace{-2mm}
    \minipage{0.32\textwidth}
        \resizebox{!}{4.74cm}{
            \input{figures/results_1.tex}
        }
    \endminipage%
    \hspace{5.5mm}
    \minipage{0.32\textwidth}%
        \resizebox{!}{4.74cm}{
            \input{figures/results_2.tex}
        }%
    \endminipage%
    \hspace{-1.5mm}
    \minipage{0.32\textwidth}%
        \resizebox{!}{4.74cm}{
            \input{figures/results_3.tex}
        }
    \endminipage
    \vspace{-4mm}
    \caption{\footnotesize Comparison of ANNs and SNNs with NF- and DFE-structure. Left: Channel A, including CDFE, LMMSE, as well as the ANN and SNN references of \cite{arnold22neuro}. 
    Middle: Channel B. Right: Channel B for networks trained at $\sigma^2=-17\,\text{dB}$ versus those trained at each $\sigma^2$.}
    \label{fig:results}
    \vspace{-3mm}
\end{figure}

\begin{wrapfigure}{l}{0.373\textwidth}
    \vspace{-11mm}
    \hspace{-2mm}
    \begin{center}
        \resizebox{!}{4.7cm}{
            \input{figures/results_length.tex}
        }
        \vspace*{-7mm}
        \caption{Channel B for different fiber lengths.}
        \label{fig:length}
    \end{center}
    \vspace*{-1.1cm}
\end{wrapfigure}
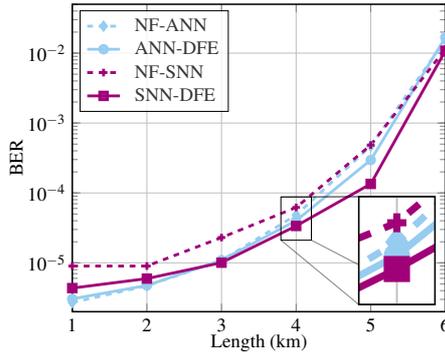
\vspace*{-2mm}

To benchmark the SNN-DFE against the approach of~\cite{arnold22neuro}, we implemented the IM/DD link as given by~\cite[Fig.~2A]{arnold22neuro} with the same parameters (channel A). 
With altered parameters, a second reference link (channel B) with a stronger CD is created.
Shared parameters are a fiber length of $\SI{5}{\kilo \meter}$, a pulse shaping roll-off factor of $\beta=0.2$, and a direct current block prior to equalization. 
Channel-specific parameters for channel A are ($100$ GBd, $1270$ nm, $-5$ $\text{ps}\,\text{nm}^{-1}\,\text{km}^{-1}$, a bias of 2.25 after pulse shaping, $n_\text{tap}=17$, $\mathcal{C}=\{-3,-1,1,3\}$) and for channel B (50 GBd, 1550 nm, $-17$ $\text{ps}\,\text{nm}^{-1}\,\text{km}^{-1}$, a bias of 0.25 after pulse shaping, $n_\text{tap}=41$, $\mathcal{C}~=~\{0,1,\sqrt{2},\sqrt{3} \} $), where $\mathcal{C}$ is the set of transmit symbols with Gray mapping.
To benchmark our SNN-DFE (using ternary encoding~\cite{Bansbach}), we implemented the SNN and reference ANN of~\cite{arnold22neuro} (using log-scale encoding) with $n_\text{tap}$ taps.
Since both approaches lack decision feedback, they are referred to as NF-SNN and NF-ANN.
Furthermore, we benchmark against the CDFE and LMMSE with $n_\text{tap}$ taps.
To compare the SNN-DFE with an ANN-based approach, we replace the SNN of~Fig.~\ref{fig:dfe} with an ANN (ANN-DFE) and ignore the encoding (real-valued inputs instead of
encoded inputs).
All networks have $N_\text{o}=4$ output neurons, for channel A $N_\text{h}=40$ hidden neurons, and channel B $N_\text{h}=80$, since the task of equalization and demapping is more demanding for strong CD. 
The networks are trained with a learning rate of $10^{-3}$ using $10\,000$ independent batches with $200\,000$ transmit symbols each.
The noise power during training is experimentally chosen to be $\sigma^2=-17\,\text{dB}$, corresponding to an ANN-DFE training SER~$\approx 10^{-2}$.
For the left plot of Fig.~\ref{fig:results}, we include the results of \cite{arnold22neuro} for their proposed SNN (ref SNN) and reference ANN (ref ANN1).
Since NF-SNN and NF-ANN outperform ref SNN and ref ANN1, which are of equal architecture, we conclude that our proposed training setup can boost performance.
As shown in the left and middle plots of Fig.~\ref{fig:results}, for both channels, the DFE-based structures outperform NF-structures; furthermore, the SNNs we studied performed better than the corresponding ANNs.
The right plot of Fig.~\ref{fig:results} reveals that training all networks at a specific noise power (t~@~17~dB) results in better performance than training the networks for each $\sigma^2$ (t~@~e~dB).
We conclude that for low~$\sigma^2$, the error magnitude is insufficient for proper learning; adequately trained networks can generalize for different~$\sigma^2$. 
Figure~\ref{fig:length} depicts the results for channel~B at $\sigma^2=21$~dB and varying fiber lengths.
Again, the networks are trained individually for each length at a $\sigma^2$ corresponding to an ANN-DFE SER~$\approx~10^{-2}$.
As the fiber length increases, the impact of CD grows, and DFE-based structures (especially the SNN-DFE) outperform NF-structures. 
We conclude that for an IM/DD link, the SNN-DFE can exploit the decision feedback to combat CD for the investigated $\sigma^2$.
It should be noted that the SNN-DFE suffers from error propagation with increasing SER.
The SNN-DFE outperforms all ANN-based equalizers and the equalizer proposed in \cite{arnold22neuro} for highly dispersive regions, enabling powerful and energy-efficient equalization and demapping. 
\vspace{-2mm}

%% file: figures/results_1.tex
\pgfplotstableread{
sigma MMSE     DFE          refANN  refSNN      ANN_MMSE        ANN_DFE         SNN_MMSE        SNN_DFE
15  0.0212725   0.01898833  0.02    0.02        1.74515508e-02  1.62663497e-02  1.59850493e-02  1.57458000e-02
16  0.01411275  0.01266333  0.012   0.012       1.05944499e-02  9.63644963e-03  9.44800023e-03  9.16939974e-03
17  0.00906     0.00809167  0.007   0.007       5.90205006e-03  5.24469977e-03  5.13949990e-03  4.93030017e-03
18  0.00546375  0.00473     0.0035  0.0035      2.98424996e-03  2.57815002e-03  2.56250007e-03  2.37015006e-03
19  0.00323     0.00286667  0.002   0.002       1.32990000e-03  1.11730001e-03  1.12775003e-03  1.02590001e-03
20  0.0018465   0.001685    0.001   0.001       5.30150020e-04  4.17100004e-04  4.43149998e-04  3.84699990e-04
21  0.001068    0.00097167  0.0005  0.0005      1.64800003e-04  1.30000000e-04  1.43450001e-04  1.18199998e-04
22  0.00054925  0.00052167  0.00025 0.0002      4.34499998e-05  3.18499988e-05  4.13000016e-05  3.12500015e-05
23  0.0002945   0.00032     0.00015 0.0000857   7.90000013e-06  7.09999995e-06  9.20000002e-06  6.14999999e-06 
}\pdpfdata

\pgfplotsset{
layers/my layer set/.define layer set={
background,
main,
up
}{
 },
    set layers=my layer set,
}

\begin{tikzpicture}[spy using outlines={rectangle, magnification=2.7, size=1cm, connect spies}]
    \def\lwidth{1.5}
    \def\opac{70}
    \def\marksz{2pt}

    \def\spywidth{2.0cm}
    \def\spyheigth{1.5cm}

    \begin{axis}[
        ymode = log,
        xscale=1,
        xlabel=$\sigma^2$ (dB),
        x label style={at={(axis description cs:0.5,0.04)},anchor=north},
        ylabel=BER,
        y label style={at={(axis description cs:0.03,0.4)},anchor=west},
        xtick = {16,18,20,22},
        xticklabels={$-16$,$-18$,$-20$,$-22$},
        grid=major,
        legend cell align={left},
        legend style={
            at={(0.02,0.01)},
            anchor=south west,
            fill opacity = 0.8,
            draw opacity = 1, 
            text opacity = 1,
        },
        xmin=15,xmax=23,
        ymin=5e-6,ymax=3e-2,
        axis line style=thick,
        tick label style={/pgf/number format/fixed},
        ]

        \coordinate (spypoint) at (axis cs:20,4.5e-4); %
        \coordinate (spyviewer) at (axis cs:21.6,7e-3); %
        \draw [fill=white] ($(spyviewer)+(1cm,0.8cm)$) rectangle ($(spyviewer)-(1cm,0.8cm)$);
        
        \spy[width=2cm,height=1.6cm, every spy on node/.append style={ultra thin},thin,line width=0.01, spy connection path={
        \draw [opacity=0.5] (tikzspyonnode.south west) -- (tikzspyinnode.south west);
        \draw [opacity=0.5] (tikzspyonnode.south east) -- (tikzspyinnode.south east);
        \draw [opacity=0.5] (tikzspyonnode.north west) -- (tikzspyinnode.north west);
        \draw [opacity=0.5] (tikzspyonnode.north east) -- (intersection of  tikzspyinnode.north east--tikzspyonnode.north east and tikzspyinnode.south east--tikzspyinnode.south west);
        ;}] on (spypoint) in node at (spyviewer); %

        \addplot[mark=none, color=black!\opac!white,line width=\lwidth,dashed, opacity= \opac] table [x = sigma, y = MMSE]{\pdpfdata};
        \addplot[mark=none,every mark/.append style={solid, fill=black!\opac!white}, color=black!\opac!white,line width=\lwidth, opacity =\opac] table [x = sigma, y = DFE]{\pdpfdata};

        \addplot[dotted,mark=star,mark size = \marksz,every mark/.append style={solid,color=KITcyanblue!\opac!white, fill=KITcyanblue!\opac!white}, color=KITcyanblue!\opac!white, line width=\lwidth] table [x=sigma, y=refANN]{\pdpfdata};
        \addplot[mark=diamond*,mark size = \marksz,every mark/.append style={solid, fill=KITcyanblue!\opac!white}, color=KITcyanblue!\opac!white, line width=\lwidth, dashed] table [x=sigma, y=ANN_MMSE]{\pdpfdata};
        \addplot[mark=*,mark size = \marksz,every mark/.append style={solid, fill=KITcyanblue!\opac!white}, color=KITcyanblue!\opac!white, line width=\lwidth] table [x=sigma, y=ANN_DFE]{\pdpfdata};

        \addplot[dotted,mark=none,mark size = \marksz, color=KITpurple,line width=\lwidth] table [x = sigma, y = refSNN]{\pdpfdata};
        \addplot[mark=+,mark size = \marksz,every mark/.append style={solid, fill=KITpurple}, color=KITpurple, dashed, line width=\lwidth] table [x=sigma, y=SNN_MMSE]{\pdpfdata};
        \addplot[mark=square*,mark size = \marksz,every mark/.append style={solid, fill=KITpurple}, color=KITpurple, line width=\lwidth] table [x=sigma, y=SNN_DFE]{\pdpfdata};

        \addlegendimage{line width=1pt, dashed, black!\opac!white} \addlegendentry{LMMSE}
        \addlegendimage{mark=triangle*,every mark/.append style={solid, fill=black!\opac!white},line width=1pt, solid, black!\opac!white} \addlegendentry{CDFE}
        
        \addlegendimage{mark=star*,every mark/.append style={solid,color=KITcyanblue!\opac!white, fill=KITcyanblue!\opac!white},line width=1pt,solid,KITcyanblue!\opac!white} \addlegendentry{ref ANN1 [4]}
        \addlegendimage{mark=diamond*,every mark/.append style={solid, fill=KITcyanblue!\opac!white},line width=1pt, dashed, KITcyanblue!\opac!white} \addlegendentry{NF-ANN}
        \addlegendimage{mark=square,every mark/.append style={fill=KITcyanblue!\opac!white,color=KITcyanblue!\opac!white},line width=1pt, dashed, KITcyanblue!\opac!white} \addlegendentry{ANN-DFE}
        
        \addlegendimage{line width=1pt,solid,KITpurple} \addlegendentry{ref SNN [4]}
        \addlegendimage{mark=+,every mark/.append style={solid, fill=KITpurple},line width=1pt, KITpurple} \addlegendentry{NF-SNN}
        \addlegendimage{mark=o,every mark/.append style={solid,fill=KITpurple},line width=1pt, dashed, KITpurple} \addlegendentry{SNN-DFE}

    \begin{pgfonlayer}{up}
    \end{pgfonlayer}
\end{axis}
\end{tikzpicture}

%% file: figures/results_2.tex
\pgfplotstableread{
sigma   ANN_MMSE        ANN_DFE         SNN_MMSE        SNN_DFE
15      2.63186991e-02  2.43450496e-02  2.22693495e-02  2.12833006e-02
16      1.69010498e-02  1.49376504e-02  1.37985498e-02  1.24191996e-02
17      1.01857996e-02  8.44809972e-03  7.99554959e-03  6.47879997e-03
18      5.61645022e-03  4.27789986e-03  4.28130012e-03  3.00185010e-03
19      2.85420008e-03  1.94244995e-03  2.07949989e-03  1.22135004e-03
20      1.28864998e-03  7.52250024e-04  9.58000019e-04  4.27999999e-04
21      5.06200013e-04  2.59599998e-04  4.12199995e-04  1.26150000e-04
22      1.77199996e-04  7.21499964e-05  1.72300002e-04  3.11000003e-05
23      5.79500011e-05  1.80999996e-05  7.62499985e-05  5.89999991e-06
}\pdpfdata

\pgfplotsset{
layers/my layer set/.define layer set={
background,
main,
up
}{
 },
    set layers=my layer set,
}

\begin{tikzpicture}[spy using outlines={rectangle, magnification=2.5, size=1cm, connect spies}]
    \def\lwidth{1.5}
    \def\opac{70}

    \def\spywidth{2.0cm}
    \def\spyheigth{1.5cm}

    \begin{axis}[
        ymode = log,
        xscale=1,
        xlabel=$\sigma^2$ (dB),
        x label style={at={(axis description cs:0.5,0.04)},anchor=north},
        yticklabels={},
        xtick = {16,18,20,22},
        xticklabels={$-16$,$-18$,$-20$,$-22$},
        grid=major,
        legend cell align={left},
        legend style={
            at={(0.02,0.01)},
            anchor=south west,
            fill opacity = 0.8,
            draw opacity = 1, 
            text opacity = 1,
        },
        xmin=15,xmax=23,
        ymin=5e-6,ymax=3e-2,
        axis line style=thick,
        tick label style={/pgf/number format/fixed},
        ]

        \coordinate (spypoint) at (axis cs:20,7e-4); %
        \coordinate (spyviewer) at (axis cs:22,3.5e-3); %
        \draw [fill=white] ($(spyviewer)+(0.7cm,1.3cm)$) rectangle ($(spyviewer)-(0.7cm,1.3cm)$);
        
        \spy[width=1.4cm,height=2.6cm, every spy on node/.append style={ultra thin},thin,line width=0.01, spy connection path={
        \draw [opacity=0.5] (tikzspyonnode.south west) -- (tikzspyinnode.south west);
        \draw [opacity=0.5] (tikzspyonnode.south east) -- (tikzspyinnode.south east);
        \draw [opacity=0.5] (tikzspyonnode.north west) -- (tikzspyinnode.north west);
        \draw [opacity=0.5] (tikzspyonnode.north east) -- (intersection of  tikzspyinnode.north east--tikzspyonnode.north east and tikzspyinnode.north west--tikzspyinnode.south west);
        ;}] on (spypoint) in node at (spyviewer); %

        \addplot[mark=diamond*,every mark/.append style={solid, fill=KITcyanblue!\opac!white}, color=KITcyanblue!\opac!white, line width=\lwidth, dashed] table [x=sigma, y=ANN_MMSE]{\pdpfdata};
        \addplot[mark=*,every mark/.append style={solid, fill=KITcyanblue!\opac!white}, color=KITcyanblue!\opac!white, line width=\lwidth] table [x=sigma, y=ANN_DFE]{\pdpfdata};

        \addplot[mark=+,every mark/.append style={solid, fill=KITpurple}, color=KITpurple, dashed, line width=\lwidth] table [x=sigma, y=SNN_MMSE]{\pdpfdata};
        \addplot[mark=square*,every mark/.append style={solid, fill=KITpurple}, color=KITpurple, line width=\lwidth] table [x=sigma, y=SNN_DFE]{\pdpfdata};

        \addlegendimage{mark=diamond*,every mark/.append style={solid, fill=KITcyanblue!\opac!white},line width=1pt, dashed, KITcyanblue!\opac!white} \addlegendentry{NF-ANN}
        \addlegendimage{mark=square,every mark/.append style={fill=KITcyanblue!\opac!white,color=KITcyanblue!\opac!white},line width=1pt, dashed, KITcyanblue!\opac!white} \addlegendentry{ANN-DFE}
        
        \addlegendimage{mark=+,every mark/.append style={solid, fill=KITpurple},line width=1pt, KITpurple} \addlegendentry{NF-SNN}
        \addlegendimage{mark=o,every mark/.append style={solid,fill=KITpurple},line width=1pt, dashed, KITpurple} \addlegendentry{SNN-DFE}

    \begin{pgfonlayer}{up}
    \end{pgfonlayer}
\end{axis}
\end{tikzpicture}

%% file: figures/results_3.tex
\pgfplotstableread{
sigma   SNN_MMSE        SNN_MMSE_e          SNN_DFE         SNN_DFE_e
15      2.22693495e-02  2.22980995e-02      2.12833006e-02  2.06211992e-02
16      1.37985498e-02  1.34347500e-02      1.24191996e-02  1.21170497e-02
17      7.99554959e-03  7.99894985e-03      6.47879997e-03  6.49964996e-03
18      4.28130012e-03  4.36295010e-03      3.00185010e-03  3.04864999e-03
19      2.07949989e-03  2.25095008e-03      1.22135004e-03  1.40904996e-03
20      9.58000019e-04  1.06529996e-03      4.27999999e-04  6.29900023e-04
21      4.12199995e-04  5.27700002e-04      1.26150000e-04  2.21099996e-04
22      1.72300002e-04  2.49600009e-04      3.11000003e-05  1.04649997e-04
23      7.62499985e-05  1.31499997e-04      5.89999991e-06  5.07500008e-05
}\pdpfdata

\pgfplotsset{
layers/my layer set/.define layer set={
background,
main,
up
}{
 },
    set layers=my layer set,
}

\begin{tikzpicture}[spy using outlines={rectangle, magnification=2.5, size=1cm, connect spies}]
    \def\lwidth{1.5}
    \def\opac{60}

    \def\spywidth{2.0cm}
    \def\spyheigth{1.5cm}

    \begin{axis}[
        ymode = log,
        xscale=1,
        xlabel=$\sigma^2$ (dB),
        x label style={at={(axis description cs:0.5,0.04)},anchor=north},
        yticklabels={},
        xtick = {16,18,20,22},
        xticklabels={$-16$,$-18$,$-20$,$-22$},
        grid=major,
        legend cell align={left},
        legend style={
            at={(0.02,0.01)},
            anchor=south west,
            fill opacity = 0.8,
            draw opacity = 1, 
            text opacity = 1,
        },
        xmin=15,xmax=23,
        ymin=5e-6,ymax=3e-2,
        axis line style=thick,
        tick label style={/pgf/number format/fixed},
        ]

        \coordinate (spypoint) at (axis cs:20,7e-4); %
        \coordinate (spyviewer) at (axis cs:21.85,4e-3); %
        \draw [fill=white] ($(spyviewer)+(0.8cm,1.2cm)$) rectangle ($(spyviewer)-(0.8cm,1.2cm)$);
        
        \spy[width=1.6cm,height=2.4cm, every spy on node/.append style={ultra thin},thin,line width=0.01, spy connection path={
        \draw [opacity=0.5] (tikzspyonnode.south west) -- (tikzspyinnode.south west);
        \draw [opacity=0.5] (tikzspyonnode.south east) -- (tikzspyinnode.south east);
        \draw [opacity=0.5] (tikzspyonnode.north west) -- (tikzspyinnode.north west);
        \draw [opacity=0.5] (tikzspyonnode.north east) -- (intersection of  tikzspyinnode.north east--tikzspyonnode.north east and tikzspyinnode.north west--tikzspyinnode.south west);
        ;}] on (spypoint) in node at (spyviewer); %

        \addplot[dashed,mark=+,every mark/.append style={solid, fill=gray!\opac!white}, color=gray!\opac!white,line width=\lwidth] table [x=sigma, y=SNN_MMSE_e]{\pdpfdata};
        \addplot[mark=+,every mark/.append style={solid, fill=KITpurple}, color=KITpurple, dashed, line width=\lwidth] table [x=sigma, y=SNN_MMSE]{\pdpfdata};

        \addplot[solid,mark=square*,every mark/.append style={solid, fill=gray!\opac!white}, color=gray!\opac!white, line width=\lwidth] table [x=sigma, y=SNN_DFE_e]{\pdpfdata};
        \addplot[mark=square*,every mark/.append style={solid, fill=KITpurple}, color=KITpurple, line width=\lwidth] table [x=sigma, y=SNN_DFE]{\pdpfdata};

        \addlegendimage{mark=+,every mark/.append style={solid, fill=gray!\opac!white},line width=1pt, gray!\opac!white} \addlegendentry{NF-SNN \;\;\;\;\,\,\ t @ $\;\,$e$\;\,\,$dB}
        \addlegendimage{mark=+,every mark/.append style={solid, fill=KITpurple},line width=1pt, KITpurple} \addlegendentry{NF-SNN \;\;\;\;\,\,\ t @ 17 dB}
        \addlegendimage{mark=o,every mark/.append style={solid,fill=gray!\opac!white},line width=1pt, dashed, gray!\opac!white} \addlegendentry{SNN-DFE $\quad\,$t @ $\;\,$e$\;\,\,$dB}
        \addlegendimage{mark=o,every mark/.append style={solid,fill=KITpurple},line width=1pt, dashed, KITpurple} \addlegendentry{SNN-DFE $\quad\,$t @ 17 dB}

    \begin{pgfonlayer}{up}
    \end{pgfonlayer}
\end{axis}
\end{tikzpicture}

%% file: figures/results_length.tex
\pgfplotstableread{
km  ANN_MMSE        SNN_MMSE        ANN_DFE         SNN_DFE
1   2.7000001e-06   9.00000032e-06  3.05000003e-06  4.34999993e-06
2   4.69999986e-06  9.00000032e-06  4.74999979e-06  5.94999983e-06
3   1.08499999e-05  2.30000005e-05  1.09000002e-05  1.00500001e-05
4   4.80999988e-05  6.19999992e-05  4.01500001e-05  3.36999983e-05
5   0.00046935      0.000484        0.00029565      0.00013465
6   0.0153555       0.011484        0.01682055      0.01081155
}\pdpfdata

\pgfplotsset{
layers/my layer set/.define layer set={
background,
main,
up
}{
 },
    set layers=my layer set,
}

\begin{tikzpicture}[spy using outlines={rectangle, magnification=2.5, size=1cm, connect spies}]
    \def\lwidth{1.5}
    \def\opac{60}

    \def\spywidth{2.0cm}
    \def\spyheigth{1.5cm}

    \begin{axis}[
        ymode = log,
        xscale=1,
        xlabel=Length (km),
        x label style={at={(axis description cs:0.5,0.04)},anchor=north},
        ylabel=BER,
        y label style={at={(axis description cs:0.03,0.42)},anchor=west},
        xtick = {1,2,3,4,5,6},
        xticklabels={$1$,$2$,$3$,$4$,$5$,$6$},
        grid=major,
        legend cell align={left},
        legend style={
            at={(0.02,0.98)},
            anchor=north west,
            fill opacity = 0.8,
            draw opacity = 1, 
            text opacity = 1,
        },
        xmin=1,xmax=6,
        ymin=2e-6,ymax=5e-2,
        axis line style=thick,
        tick label style={/pgf/number format/fixed},
        ]

        \coordinate (spypoint) at (axis cs:4,4.3e-5); %
        \coordinate (spyviewer) at (axis cs:5.35,1.5e-5); %
        \draw [fill=white] ($(spyviewer)+(0.7cm,1cm)$) rectangle ($(spyviewer)-(0.7cm,1cm)$);
        
        \spy[width=1.4cm,height=2cm, every spy on node/.append style={ultra thin},thin,line width=0.01, spy connection path={
        \draw [opacity=0.5] (tikzspyonnode.south west) -- (tikzspyinnode.south west);
        \draw [opacity=0.5] (tikzspyonnode.south east) -- (intersection of  tikzspyinnode.south east--tikzspyonnode.south east and tikzspyinnode.north west--tikzspyinnode.south west);
        \draw [opacity=0.5] (tikzspyonnode.north west) -- (tikzspyinnode.north west);
        \draw [opacity=0.5] (tikzspyonnode.north east) -- (tikzspyinnode.north east);
        ;}] on (spypoint) in node at (spyviewer); %

        \addplot[mark=diamond*,every mark/.append style={solid, fill=KITcyanblue!\opac!white}, color=KITcyanblue!\opac!white, line width=\lwidth, dashed] table [x=km, y=ANN_MMSE]{\pdpfdata};
        \addplot[mark=*,every mark/.append style={solid, fill=KITcyanblue!\opac!white}, color=KITcyanblue!\opac!white, line width=\lwidth] table [x=km, y=ANN_DFE]{\pdpfdata};

        \addplot[mark=+,every mark/.append style={solid, fill=KITpurple}, color=KITpurple, dashed, line width=\lwidth] table [x=km, y=SNN_MMSE]{\pdpfdata};
        \addplot[mark=square*,every mark/.append style={solid, fill=KITpurple}, color=KITpurple, line width=\lwidth] table [x=km, y=SNN_DFE]{\pdpfdata};

        \addlegendimage{mark=diamond*,every mark/.append style={solid, fill=KITcyanblue!\opac!white},line width=1pt, dashed, KITcyanblue!\opac!white} \addlegendentry{NF-ANN}
        \addlegendimage{mark=square,every mark/.append style={fill=KITcyanblue!\opac!white,color=KITcyanblue!\opac!white},line width=1pt, dashed, KITcyanblue!\opac!white} \addlegendentry{ANN-DFE}
        
        \addlegendimage{mark=+,every mark/.append style={solid, fill=KITpurple},line width=1pt, KITpurple} \addlegendentry{NF-SNN}
        \addlegendimage{mark=o,every mark/.append style={solid,fill=KITpurple},line width=1pt, dashed, KITpurple} \addlegendentry{SNN-DFE}

\end{axis}
\end{tikzpicture}

%% file: imdd_dfe.bbl
\begin{thebibliography}{99} %
\vspace*{-1mm}
\bibitem{Ferreira} \footnotesize T. Ferreira de Lima \textit{et al.} ``Progress in neuromorphic photonics,'' \textit{Nanophotonics}, vol. 6, no. 3, pp. 577-599, 2017.

\bibitem{BPTT} \footnotesize E. O. Neftci, H. Mostafa and F. Zenke, ``Surrogate gradient learning in spiking neural networks: Bringing the power of gradient-based optimization to spiking neural networks,''
\textit{IEEE Signal Process. Mag.}, vol. 36, no. 6, pp. 51-63, Nov. 2019

\bibitem{Bansbach} \footnotesize E.-M. Bansbach, A. von Bank and L. Schmalen, ``Spiking neural network decision feedback equalization'', in \textit{Proc. Int. ITG WSA-SCC}, Braunschweig, Germany, Feb. 2023

\bibitem{arnold22neuro} \footnotesize E. Arnold \textit{et al.}, ``Spiking neural network equalization on neuromorphic hardware for IM/DD optical communication,'' in \textit{Proc. Eur. Conf. Opt. Commun. (ECOC)}, Basel, CH, Sep. 2022.

\bibitem{norse} \footnotesize C. Pehle and J. Pedersen, ``Norse -- A deep learning library for spiking neural networks,'' Jan. 2021, doi:10.5281/zenodo.4422025, Documentation: https://norse.ai/docs/







\end{thebibliography}
